\documentclass[preprint,12pt]{elsarticle}

\usepackage{amssymb}
\usepackage{amsmath}
\usepackage{amsthm}

\usepackage{subcaption}
\usepackage{algorithm}
\usepackage{algorithmic}

\usepackage{natbib}
\biboptions{sort&compress}

\journal{Franklin Open}

\DeclareMathOperator*{\argmax}{arg\,max}

\begin{document}

\begin{frontmatter}



\title{Belief States for Cooperative Multi-Agent Reinforcement Learning under Partial Observability}


\author{Paul J. Pritz, Kin K. Leung} 

\affiliation{organization={Department of Computing, Imperial College London},
            city={London},
            country={United Kingdom}}

\begin{abstract}
Reinforcement learning in partially observable environments is typically challenging, as it requires agents to learn an estimate of the underlying system state.
These challenges are exacerbated in multi-agent settings, where agents learn simultaneously and influence the underlying state as well as each others' observations.
We propose the use of learned beliefs on the underlying state of the system to overcome these challenges and enable reinforcement learning with fully decentralized training and execution.
Our approach leverages state information to pre-train a probabilistic belief model in a self-supervised fashion.
The resulting belief states, which capture both inferred state information as well as uncertainty over this information, are then used in a state-based reinforcement learning algorithm to create an end-to-end model for cooperative multi-agent reinforcement learning under partial observability.
By separating the belief and reinforcement learning tasks, we are able to significantly simplify the policy and value function learning tasks and improve both the convergence speed and the final performance.
We evaluate our proposed method on diverse partially observable multi-agent tasks designed to exhibit different variants of partial observability.
\end{abstract}



\begin{keyword}
multi-agent reinforcement learning \sep partial observability \sep belief states \sep decentralized training and decentralized execution


\end{keyword}

\end{frontmatter}
\noindent
Funding: This work was supported by the Dstl SDS Continuation project and EPSRC grant EP/Y037243/1.


\section{Introduction}

Many real-world applications for reinforcement learning (RL) inherently involve multi-agent systems and frequently suffer from incomplete state information, i.e., partial visibility of the underlying system state.
Prominent examples of such applications include drone management, object search, and traffic control \cite{pham2018cooperative,kong2017collaborative,calvo2018heterogeneous}.
Traditional approaches in the multi-agent domain attempt to solve this by introducing memory via recurrent networks in the policy and value function architecture and by allowing communication or by assuming the central availability of all data during training.
All of these approaches have significant drawbacks. 
First, state representations are learned solely using the reward as a signal, which is not predictive of underlying state features or environment dynamics.
Second, there is no notion of uncertainty included in these representations.
Third, assuming central data availability or allowing communication may be an unrealistic assumption in many settings or require significant communication overhead.
The assumption of centralized data availability during the training phase gives rise to the centralized training, decentralized execution paradigm (CTDE).
Typically, a centralized critic would process all agent observations or even the underlying state and then provide a learning signal for individual agent policies \cite{lowe2017multi}.
Although this allows agents to act in a decentralized way, it places strong restrictions on training, assuming that all information is available to a central critic.

We propose a framework that does not rely on this assumption for policy learning and efficiently addresses partial observability.
Specifically, we propose a probabilistic model to infer the underlying system states from local agent histories.
This \textit{belief model} uses conditional variational auto-encoders \cite{kingmaauto} to be able to infer states and quantify the uncertainty associated with the prediction.
Assuming a small amount of transition data collected from the environment, our belief model can be pre-trained in a self-supervised fashion and then used in a downstream RL task.
We assume that the full state information required for the self-supervised training of the belief model is available from the environment during this pre-training phase but not during the following reinforcement learning phase.
By separating the representation learning task from the reinforcement learning task through the trained belief model, we are able to use a better prediction target, i.e., states rather than rewards, for the representation learning aspect and still facilitate fully decentralized reinforcement learning. 
For the exact model specification, we build on recent advances from the single-agent domain \cite{wang2023learning} and adapt these to multi-agent settings.

To use the proposed belief model in a multi-agent RL setting, we present an extension of the I2Q model \cite{jiang2022i2q}, incorporating the learned belief states to train value functions per agent.
I2Q has been theoretically shown to converge in cooperative settings, even when agents are simultaneously trained in a decentralized way.
Our proposed approach, therefore, follows the decentralized training, decentralized execution paradigm (DTDE) and does not rely on any communication among agents during the RL part.

In order to evaluate our approach, we propose a set of partially observable multi-agent domains designed to restrict state observability in different ways.
Specifically, we consider scenarios with information asymmetry, coordination requirements, and memory requirements.

\section{Related Work}
Multi-agent reinforcement learning (MARL) can be broadly categorized into centralized training, decentralized execution (CTDE) and decentralized training, decentralized execution (DTDE) approaches \cite{gronauer2022multi}.
In the former, agents are updated using mutual information (such as a shared state), which is then discarded at execution (test) time, allowing each agent to act on local information.
The latter approach, restricts the information available to agents to local observations, both during training and execution.
In addition to these MARL paradigms, multi-agent adaptations of single-agent RL algorithms still remain popular due to their simplicity and encouraging performance in some domains \cite{tan1993multi,yu2022surprising}.
In the following, we focus on recent work relevant to the partially observable DTDE setting, i.e., work trying to alleviate the pathologies arising from non-stationarity as well as partial observability in MARL \cite{gronauer2022multi}. 

\subsection{Non-stationarity in MARL}
Multiple agents learning simultaneously leads to the problem of non-stationarity.
If each agent treats the other agents as part of the environment, the transition function from the individual agent's perspective changes over time as the other agents' policies evolve, leading to a non-stationary environment \cite{oliehoek2016concise}.
This \textit{moving-target} problem also means that convergence guarantees for single-agent RL algorithms no longer hold in multi-agent scenarios with decentralized training.
CTDE has, therefore, emerged as a popular paradigm, allowing agents to share information during training while maintaining independent execution \cite{gronauer2022multi}.
Several pieces of previous research propose algorithms where a centralized critic with full visibility provides a learning signal to decentralized actors \cite{lowe2017multi,iqbal2019actor,peng2021facmac,rashid2018qmix,son2019qtran,foerster2018counterfactual}. 
While such centralized training approaches address the non-stationarity problem through joint, centralized training, they also come with restrictive assumptions on state and action visibility.
On the other end of the spectrum, methods following a DTDE approach must explicitly deal with non-stationarity.
To this end, \citet{matignon2007hysteretic,jiang2022i2q,lauer2000algorithm,jiang2023best} propose Q-learning variants, adapted for multi-agent learning, which come with convergence guarantees in cooperative settings. 
\citet{jiang2022i2q} develop a method which is based on state-state Q-values while the others use modified update functions for the state-action Q-values.
Other approaches adapt single-agent policy-based methods for multi-agent domains \cite{samvelyan2019starcraft,su2022decentralized}.
The gap between centralized and decentralized training schemes is bridged by methods that maintain estimates of other agents' policies or share partial information via communication protocols \cite{oroojlooy2023review,zhang2021multi}.
Typically, agents learn to communicate either state or policy-related information \cite{sukhbaatar2016learning,singh2018learning,foerster2016learning} or to predict other agent's policies \cite{zhai2023dynamic,grover2018learning,hong2017deep,papoudakis2021agent}.

\subsection{Partial Observability}
Partial observability presents significant challenges in both single-agent and multi-agent reinforcement learning settings.
In single-agent RL, the most common methods for learning in partially observable environments are based on the use of recurrent networks to process a history of observations \cite{karkus2017qmdp}.
More recently, probabilistic methods and the use of belief models based on variational objectives have gained traction \cite{wang2023learning,guo2018neural,venkatraman2017predictive}. 
These models typically encode beliefs by learning a distribution, conditioned on the observation history, that is predictive of future states or rewards.
In MARL, the challenges induced by partial observability are further exacerbated by potential information asymmetries between agents \cite{wong2023deep}.
The theoretical framework for partially observable MARL settings is either a partially observable Markov game for mixed and competitive settings or a decentralized partially observable Markov decision process (Dec-POMDP) for cooperative scenarios \cite{bernstein2002complexity}.

Since we restrict ourselves to cooperative settings, our analysis will focus on research in this area.
Several pieces of previous research propose applying algorithms designed for single-agent RL with partial observability to the multi-agent domain.
\citet{gupta2017cooperative} propose such an extension of policy gradient-based RL to multi-agent problems with centralized training.
Following a similar idea, \citet{wang2020r} investigate a recurrent version of MADDPG proposed by \citet{lowe2017multi} to deal with partial observability at the centralized critic level.
\citet{diallo2019} and \citet{park2020cooperative} also present CTDE methods to address partial observability.
While \citet{diallo2019} rely on a simple central Q-value critic, \citet{park2020cooperative} propose a communication framework between agents, which is also centrally trained.

A further strand of MARL research is concerned with the use of learned belief states or embeddings to address partial observability in the MARL domain.
\citet{mao2020information} present a model where an embedded version of the environment state is learned via recurrent networks and then used in the agents' policy.
Other approaches utilize variational models to either learn beliefs over environment states in a CTDE setting \cite{zhang2022common} or to learn beliefs over other agents' observations and policies \cite{wen2019probabilistic,moreno2021neural} 
Overall, few approaches consider partial observability in a DTDE setting, and the existing methods frequently rely on variants of single-agent algorithms.

\section{Methodology}
Inspired by belief state-based approaches in single-agent RL, we propose a representation learning module for partially observable multi-agent problems \cite{wang2023learning}.
Our proposed approach is coupled with state-based Q-learning \cite{jiang2022i2q} and can be used in fully decentralized multi-agent reinforcement learning.
Based on local agent histories, our goal is to learn probabilistic beliefs over the underlying system state that serve as representations for the downstream RL task.
Using these beliefs, we train a set of policies  $\pi_i(a_i | o_i, b(h_i))$, where $o_i$ is the current observation and $b(h_i)$ is the belief over the system state given the history $h_i$ for agent $i$.
To this end, we propose a two-stage learning process.
First, we pre-train a belief model for each agent, assuming access to an offline labelled dataset collected from the environment.
Second, belief states are used in a variant of the I2Q framework to learn policies in a decentralized fashion with access to local information only.

\subsection{Preliminaries}
In this work, we restrict ourselves to cooperative multi-agent scenarios with partial observability.
The environment can, therefore, be defined as a decentralized, partially observable Markov decision process (Dec-POMDP) \cite{bernstein2002complexity}
A Dec-POMDP is described by a tuple $\langle S, O, A, R, P, \mathcal{W}, \gamma \rangle$.  In this framework, $o_i \in O$ represents the partial observation available to an agent, derived from the global system state $s \in S$, while $a_i \in A$ denotes the action taken by agent $i$. Each agent $i \in N:= \{1, \dots, N\}$ has access to a partial observation $o_i$, which is determined through the observation function $\mathcal{W}(s): S \rightarrow O$.
The combined actions of all $N$ agents are denoted by the joint action $\textbf{a}:= a_{i \in N} \in A^N$. The state transitions follow the probability distribution $P(s'|s, \textbf{a}) : S \times A^N \times S \rightarrow [0,1]$. At each step, agents observe a joint reward as per the reward function $R(s, s'): S \times S \rightarrow \mathbb{R}$. 
The goal of the agents is to learn a joint policy $\boldsymbol\pi = \prod_i \pi_i(a_i | o_i)$ that maximizes the joint return $G = \sum_t \gamma^t r_t$, where $G$ is the sum of expected discounted future rewards at time step $t$ and $\gamma$ is the discount factor.

To train a set of cooperative agents in a decentralized fashion, we build on a recent MARL approach called I2Q, proposed by \citet{jiang2022i2q}, which addresses the non-stationarity issue in cooperative settings by extending the work of \citet{edwards2020qss}.
We briefly introduce their approach in the following and refer the reader to \citet{jiang2022i2q} for a more detailed theoretical analysis.
\citet{jiang2022i2q} show that agents performing Q-learning on the ideal transition probabilities $P^*_i(s' | s, a_i) = P(s' | s, a_i, \pi^*_{-i}(s, a_i))$ in deterministic environments will converge to their individual optimal policy.
Ideal transition probabilities assume optimal behaviour by all other agents and the existence of an optimal joint policy $\pi^*(s) = \argmax_a Q(s, \textbf{a})$.
Since the optimal policy and transition probabilities are not known a priori, a state-state value function $Q^{ss}_i(s, s')$ is learned for each agent
\[
    Q^{ss}_i(s, s') = r + \gamma \max_{s'' \in \mathcal{N}(s')} Q^{ss}_i(s', s'') \,,
\]
where $\mathcal{N}(s')$ is the neighbouring set of next states.
This function can be learned from local non-stationary experiences and learns values equivalent to the joint Q-function $\max_{s'} Q_i^{ss}(s, s') = \max_{a} Q(s, \textbf{a}).$
To avoid direct maximization over neighbouring states, which can be unfeasible in large or continuous state environments, each agent subsequently learns a transition function $f_i(s, a_i)$ by maximizing the expectation
\[
    \mathbb{E}_{s, a_i, s'} \left[ \lambda Q^{ss}_i(s, f_i(s, a_i)) - (f_i(s, a_i) - s')^2 \right] \,.
\]
Here, the first term encourages the next states with high $Q^{ss}$ values, and the second term constrains them to the set of neighbouring states.
A state-action Q-function can then be learned using $Q^{ss}$ and $f_i$ by minimizing
\[
    \mathbb{E}_{s, a_i, r \sim \mathcal{D}_i} \left[ \left( Q_i(s, a_i) - r - \gamma \max_{a'_i} \bar{Q}_i(f_i(s, a_i), a'_i) \right)^2 \right] \,.
\]
While this approach allows fully decentralized training and execution, \citet{jiang2022i2q} make several assumptions about the environment.
They assume that the environment is deterministic and cooperative and that all agents share and observe the same state.
In this work, we relax the assumption about shared states and instead assume a partially visible environment where the true underlying state is not observed during reinforcement learning. 
The other assumptions remain the same for our approach.

\subsection{Self-Supervised Learning of Belief States}
\label{sec:belief_model}

\begin{figure}
    \centering
    \includegraphics[width=0.8\linewidth]{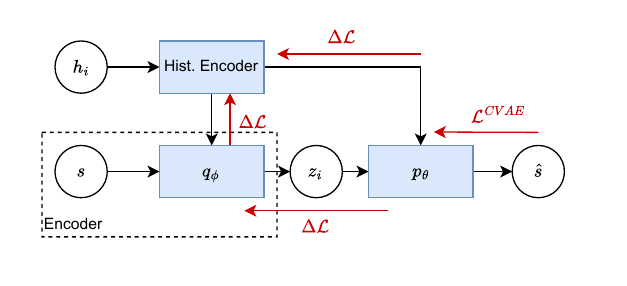}
    \caption{CVAE architecture to learn beliefs over states from local histories. The encoder part of the model is only used during pre-training and discarded for the RL part. All model components are trained using the $\mathcal{L}^{CVAE}$ loss. Both encoder and decoder losses are back-propagated to the history encoder. Each agent maintains its own model.}
    \label{fig:belief_model}
\end{figure}
To model beliefs over the underlying system state, we propose using a conditional variational auto-encoder (CVAE) \cite{kingmaauto} on a per-agent level.
Training an individual model per agent not only allows us to keep the RL training fully decentralized but also allows us to address potential information asymmetries between agents.
For the downstream RL task, we want to find a distribution $p(s|h_i)$ for each agent that estimates the system state from the agent's local history.
For learning these belief states, we assume we have access to a pre-collected dataset from the environment,  $(s, \{h_i\}_{i\in N}) \sim \mathcal{D}$, where $h_i$ is agent $i$'s history of observations and actions up to the current time step. Note that we omit the time step subscript in all our formulas for ease of exposition.
We further assume that for a given history $h_i$, observing state $s$ is governed by a latent variable $z_i$ \cite{wang2023learning}.
The underlying system state can hence be modelled as a stochastic function, conditioned on $h_i$ and $z_i$
\[
p(s, h_i, z_i) = p(h_i)p(z_i)p(s| h_i, z_i)\,.
\]
Based on this generative model, we define an encoder-decoder architecture to learn the distribution over the unobserved system state.
This is comprised of an encoder to approximate the posterior distribution $p(z_i| s, h_i)$ as $q_{\phi}(z_i | s, h_i)$ and a decoder $p_{\theta}(s | h_i, z_i)$, parametrized by $\phi$ and $\theta$.
To train the CVAE, we maximize the conditional log-likelihood of observed data $\log p_\theta(s| h_i)$, using the evidence lower bound (ELBO) \cite{kingmaauto}.
This is defined as
\begin{align}
    \log p_{\theta}(s | h_i) &\geq \mathbb{E}_{q_{\phi}(z_i | s, h_i)} \left[\log p_{\theta}(s | h_i, z_i)  + \log p(z_i) - \log q_{\phi}(z_i | s, h_i) \right]. 
\end{align}
Splitting the terms, we get
\begin{align}
    \log p_{\theta}(s | h_i) &\geq \mathbb{E}_{q_{\phi}(z_i | s, h_i)} \left[\log p_{\theta}(s | h_i, z_i) \right] \\
    & - \mathbb{E}_{q_{\phi}(z_i | s, h_i)} \left[\log p(z_i) - \log q_{\phi}(z_i | s, h_i) \right]. \nonumber
\end{align}
The second expectation on the right-hand side above is the KL divergence between the prior and learned posterior.
Our loss function for the CVAE model therefore becomes
\begin{align}
    \mathcal{L}^{CVAE}_i(s | h_i) &= \mathbb{E}_{q_{\phi}(z_i | s, h_i)} \left[\log p_{\theta}(s | h_i, z_i)\right] - D_{KL}(q_{\phi}(z_i | s, h_i) || p(z_i)).
    \label{eq:loss_belief}
\end{align}
Note that in practice, we can instantiate a single version of this model, which is then shared among agents after pre-training if all agents have the same observations.
In the case where observations differ between agents, one model per agent is required.
Both the encoder and decoder are parametrized by neural networks, and we use a recurrent history encoder, which is shared by the encoder and decoders.
An overview of this model can be seen in Figure \ref{fig:belief_model}.

\begin{algorithm}
\caption{Belief Model Pre-training}
\begin{algorithmic}
\STATE \textbf{Input:} Dataset $\mathcal{D}_i = \{s, h^i\}$ 
\WHILE{not converged}
    \STATE Sample $s, h_i \sim \mathcal{D}_i$
    \STATE $\mathcal{L}^{CVAE}_i(s | h_i) = \mathbb{E}_{q_{\phi}(z_i | s, h_i)} \left[\log p_{\theta}(s | h_i, z_i)\right] - D_{KL}(q_{\phi}(z_i | s, h_i) || p(z_i))$
    \STATE Update $\phi, \theta$ and history encoder using $\mathcal{L}^{CVAE}$
\ENDWHILE
\end{algorithmic}
\end{algorithm}

\subsection{Learning the Value Function~--~Belief-I2Q}
We conduct training by using local observations combined with the previously learned belief state.
As the decoder of the CVAE outputs a distribution over states, we represent this by a collection of i.i.d. samples from the decoder's distribution over states.
For each agent, we draw $m$ samples from the prior $p(z_i)$ to use in the decoder network to obtain $\hat{s_j} \sim p_{\theta}(s_j|h_{i,j}, z_{i,j}) \forall j<m$.
These samples include both the mean and variance and are then averaged such that 
\begin{equation}
    b(h_i) = \frac{1}{m} \sum_j \hat{s_j} \,.
\end{equation}
For ease of notation, we define the encoding $g_i = g(o_i, b(h_i))$ as the concatenated observation and belief state for agent $i$.
To avoid the credit assignment problem faced by the state-state value function $Q^{ss}$, caused by multiple observations potentially mapping to the same state, we leave this in observation space.
Therefore, each agent learns three functions: (1) $Q^{ss}_i$, (2) $f_i(g_i,a_i)$, (3) $Q_i(g_i, a_i)$.
The state-state value function is trained to minimize the objective,
\begin{multline}
\mathcal{L}^{Q^{ss}}_i = \mathbb{E}_{o_i,a_i,o'_i,r \sim \mathcal{D}_i} \left[ \left( Q^{ss}_i(o_i, o'_i) - r - \gamma \bar{Q}^{ss}_i(o'_i, f_i(g'_i, a^*_i)) \right)^2 \right], \\
\quad a^*_i = \arg\max_{a'_i} Q_i(g'_i, a'_i).
\label{eq:loss_qss}
\end{multline}
The transition function $f_i$ is learned by maximizing
\begin{equation}
    \mathcal{L}^f_i = \mathbb{E}_{o_i,a_i,o'_i \sim \mathcal{D}_i} \left[ \lambda Q^{ss}_i(o_i, f_i(g_i, a_i)) - (f_i(g_i, a_i) - o'_i)^2 \right].
    \label{eq:loss_f}
\end{equation}
As we still require a state-action value function to assess actions executed in the environment, we further train the Q-function $Q_i(g_i, a_i)$ for each agent.
This value function $Q_i(g_i, a_i)$ is learned by minimizing
\begin{equation}
    \mathcal{L}^Q_i = 
    \mathbb{E}_{o_i, a_i, r \sim \mathcal{D}_i} \left[ \left( Q_i(g_i, a_i) - r - \gamma \max_{a'_i} \bar{Q}_i(g(f_i(g_i, a_i), b(h_i')), a'_i) \right)^2 \right] \,,
    \label{eq:loss_q}
\end{equation}
where $b(h')$ is the belief state given the agent's history extended by the next observation predicted by $f_i$, and $\bar{Q}_i$ is the target network of $Q_i$.
All three functions (1) - (3) are updated iteratively from experiences collected in the environment.
Our setup to learn the value function is an extension of I2Q presented by \citet{jiang2022i2q} and we refer to our method as \textbf{Belief-I2Q}.

\subsection{Architecture and Training}
We implement all functions as neural networks and conduct training as a two-stage process.
First, the agents gather a small amount of data from the environment using a random roll-out policy.
During this period, we assume that agents have access to the true state information, which is used to label the episode data.
To learn the belief state model outlined in Section \ref{sec:belief_model}, we use recurrent history encoders to process agent history $h_i$.
The encoding is then used as input to both the encoder and decoder networks $p_\theta$ and $q_\phi$, and we jointly train these to minimize \eqref{eq:loss_belief}.

After pre-training of the belief model, all agent networks are initialized and trained in a fully decentralized fashion.
The Q-functions and transition functions are then updated iteratively from experience collected in the environment using losses \eqref{eq:loss_qss} - \eqref{eq:loss_q}.
Importantly, this stage of the training process does not assume access to any underlying system states.

\section{Experiments}

\subsection{Environments}
To evaluate the effectiveness of our approach in partially observable MARL settings, we construct a set of grid world environments.
As per the requirements of state-based Q-learning, all studied scenarios are cooperative, i.e., agents share the same reward.
The environments we use are designed to test the effectiveness of our approach under different types of partial observability, e.g. asymmetric information, local visibility, and temporally restricted visibility.

\begin{figure}[htb]
    \centering
    \begin{subfigure}[t]{0.45\textwidth}
        \includegraphics[width=\textwidth]{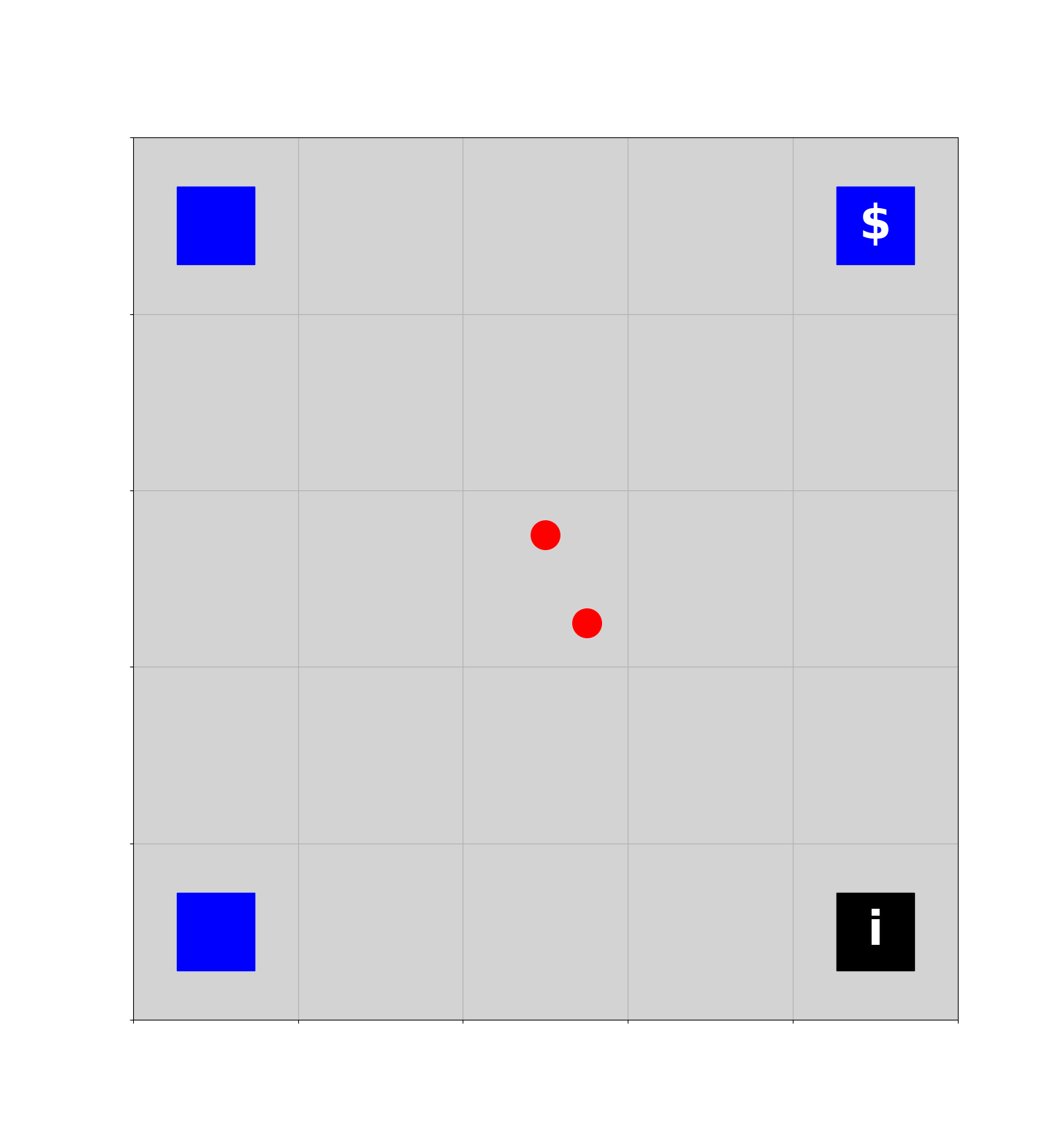}
        \caption{Visualization of the Oracle environment. The red dots represent the agents (starting from the centre of the grid).
        Blue boxes represent treasure locations, and the black box represents the Oracle. Only one treasure location contains a reward.}
        \label{fig:oracle}
    \end{subfigure}
    \hfill
    \begin{subfigure}[t]{0.45\textwidth}
        \includegraphics[width=\textwidth]{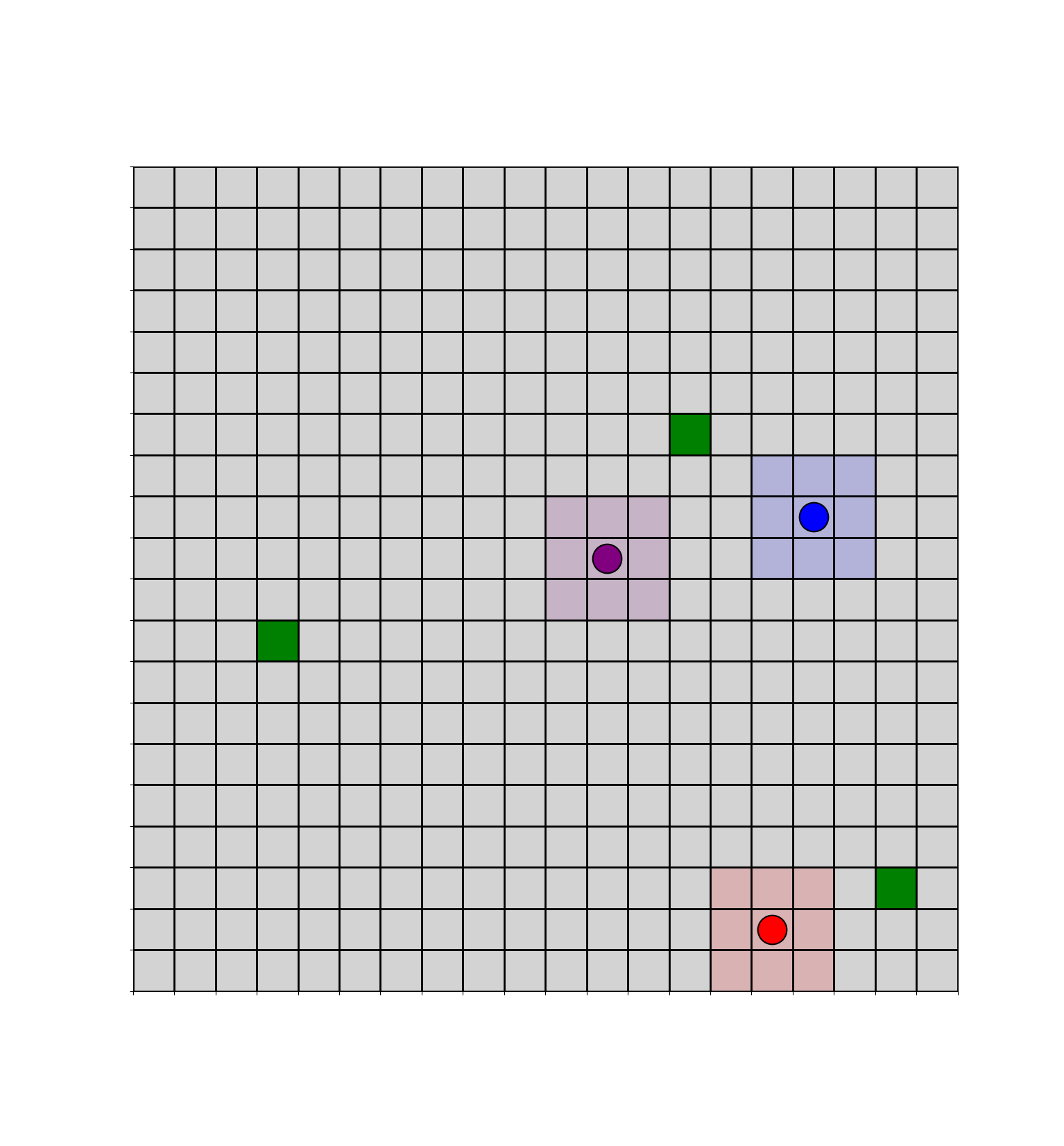}
        \caption{Visualization of the Gathering environment. Green boxes represent randomly placed reward fields. The coloured dots are agents, starting in random positions on the grid. Their visibility radius is indicated by the shaded area around them.}
        \label{fig:gathering}
    \end{subfigure}
    
    \caption{Oracle and Gathering environments.}
    \label{fig:envs_1}
\end{figure}

\begin{figure}[htb]
    \begin{subfigure}[t]{0.45\textwidth}
        \includegraphics[width=\textwidth]{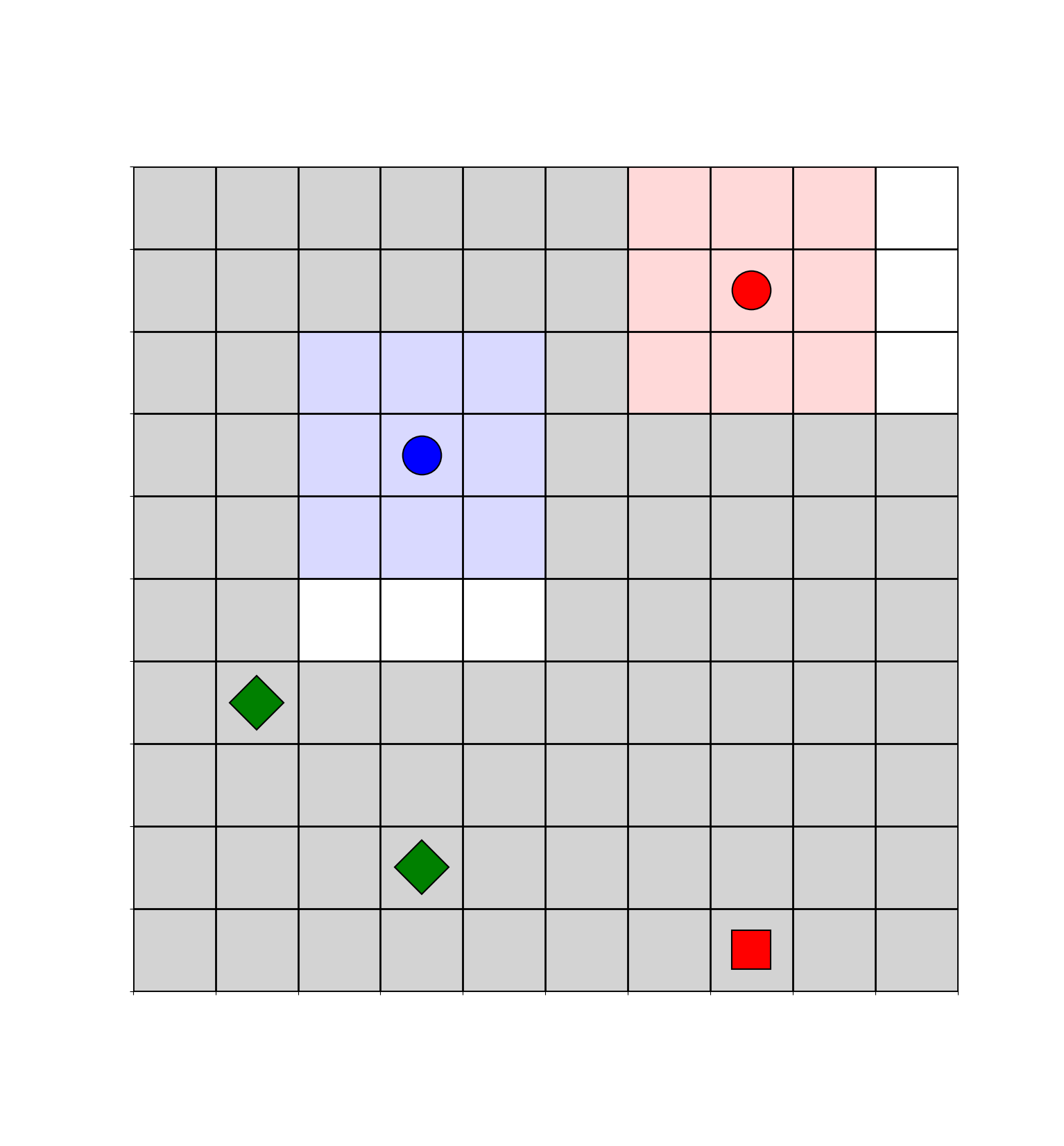}
        \caption{Visualization of the Escape Room environment. Agents are randomly placed on the grid at the start of an episode. Their visibility radius is indicated by the shaded area around them. The keys are symbolized by the green rhombuses, while the exit is represented by a red square. Already explored areas are marked as white, while the unexplored areas are grey.}
        \label{fig:escape}
    \end{subfigure}
    \hfill
    \begin{subfigure}[t]{0.43\textwidth}
        \includegraphics[width=\textwidth]{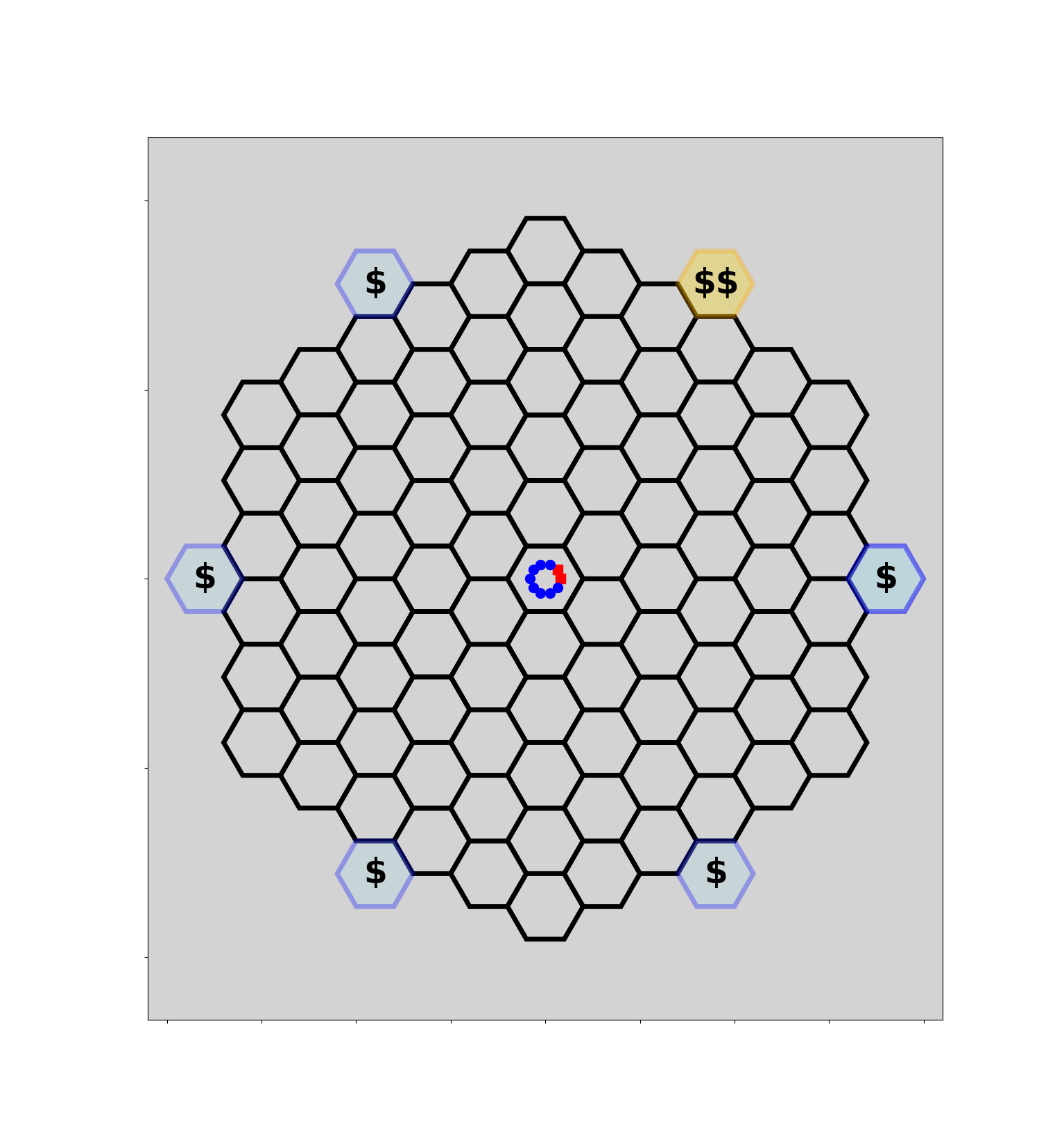}
        \caption{Visualization of the Honeycomb environment. Two \textit{informed} agents (red) and 8 \textit{uninformed} agents (blue) are placed in the centre of the field. The informed agents are aware of the higher reward field (orange with \$\$ sign). The uninformed agents see all six reward fields as equal value.}
        \label{fig:honeycomb}
    \end{subfigure}
    \caption{Escape Room and HoneyComb environments.}
    \label{fig:envs_2}
\end{figure}

\subsubsection{Scenario 1: Multi-Agent Oracle}
For this environment, we adopt the information-seeking problem proposed by \citet{wang2023learning} to a multi-agent setting.
The agents are randomly placed in a grid world, with treasures placed at three corners of the grid.
Only one randomly chosen treasure yields a positive reward.
Steps are penalized with a small step penalty.
The location of the treasure containing the reward is unknown to the agents but can be queried from an \textit{oracle} positioned in the remaining corner of the grid.
Once queried, the location of the higher reward field is revealed for one time step.
All agents share the same observation and have full visibility of the grid apart from the correct treasure location.
The state always contains the correct treasure location.
Observations are continuous and consist of both agents' coordinates as well as the location of the correct treasure once revealed.
There are 9 discrete actions, with 8 actions that move the agent in different directions and one void action.
An illustration can be seen in Figure \ref{fig:oracle}.

\subsubsection{Scenario 2: Gathering Task}
Agents are again randomly placed in a grid world, and their task is to gather a number of equally valuable treasures scattered around the grid.
All treasures yield an equal positive reward, while each step has a small cost associated with it. 
Each agent has a local visibility radius.
The parts of the grid visible to the agents at any given time step are combined to create the observation.
All areas of the map, currently not in any agents' radius are not visible.
An illustration of this environment can be seen in Figure \ref{fig:gathering}.
Agents and rewards are encoded as indices, and observations and states are discrete.
The state consists of the grid with full visibility.
Five discrete actions are available to the agents, four of which move them to the adjacent fields and one being the void action.

\subsubsection{Scenario 3: Multi-Agent Escape Room}
The escape room scenario requires agents to collect a number of keys randomly scattered around a grid to be able to unlock the exit also randomly placed in the grid.
Keys can be collected by any agent and the exit can be unlocked once they jointly hold all keys.
Only unlocking the exit yields a positive reward, whereas steps and collisions between the agents yield a negative reward.
As agents move around the grid, they gradually uncover it with their visibility radius.
The observation contains all the explored areas and is not obfuscated again once agents move away from a discovered area.
The observation contains the entire grid, with integer representations for the different items and default values for the undiscovered areas.
A visualization of this environment can be found in Figure \ref{fig:escape}.
The state contains the fully revealed grid.
Again, there are 5 discrete actions available, including moves to the adjacent fields and a void action.

\subsubsection{Scenario 4: Coordinated HoneyComb}
This setting, originally proposed by \citet{boos2019honeycomb}, places agents in the centre of a hexagonal (honeycomb-like) field.
Six corners of the field contain a reward field.
Two of these reward fields yield a higher reward, while all others yield an equal positive reward.
Additionally, there is a multiplicative bonus if multiple agents reach the same reward field.
Information asymmetry introduces partial observability: only two of the 10 agents (\textit{informed}) know the location of the higher-reward fields.
Therefore, their task becomes to establish a leader-follower dynamic, guiding the uninformed agents to the higher-reward fields \cite{boos2014leadership}.
Observations are represented by the coordinate positions of all agents and the location of the reward fields.
Informed agents additionally observe the locations of the two higher-reward fields, these are also included in the state.
The 7 available actions are discrete and include moving to the adjacent hexagons and a void action.
This environment is visualized in Figure \ref{fig:honeycomb}.

\subsection{Experimental Setup and Benchmarks}
We compare our approach \textit{Belief-I2Q} against two approaches that follow the DTDE paradigm.
Our first benchmark is a recurrent version of the I2Q algorithm (\textit{Rec.-I2Q}) \cite{jiang2022i2q}.
To endow the method with memory and enable learning in the partially observable environments we consider, we implement a per-agent history encoder, which is shared between the $Q^{ss}_i$ and the $Q_i$ function for each agent. 
Our second benchmark is a recurrent version of hysteretic independent Q-learning (\textit{Rec.-Hyst.-IQL}) \cite{matignon2007hysteretic}, again using a per-agent history encoder, which is used by the Q-function.
In contrast to independent Q-learning, hysteretic Q-learning applies different learning rates to positive and negative experiences.
The idea is to associate less weight with experiences that resulted in low rewards, potentially caused by the behaviour of other agents.
The modified Q-learning update for hysteretic Q-learning is $Q_i(o_i, a_i) \leftarrow Q_i(o_i, a_i) + \alpha \psi$ for $\psi > 0$ and $Q_i(o_i, a_i) \leftarrow Q_i(o_i, a_i) + \beta \psi$ otherwise, where $\psi = r + \gamma \max_{a'} Q_i(o'_i, a') - Q_i(o_i, a_i)$ and $\alpha > \beta$. 
Both baseline approaches do not leverage separately learned state representations but instead follow the traditional approach of using recurrent networks to learn from observation histories.
A further key difference between the baselines and our approach is that the baseline algorithms learn representations of the environment using only the reinforcement learning loss as they train their history encoders as part of the Q-functions.
As all our environments have discrete actions, we use the learned Q-functions directly as an action policy.
Complete parametrization of the respective architectures can be found in \ref{app:architecture}.
For continuous action environments, the approach could easily be extended using an actor function.

\subsection{Results}
To ensure comparability, we train all models for an equal number of episodes and update parameters after each episode.
We conduct a grid search over key hyperparameters and report results for the models that performed best across this search.
All results are averaged over three random seeds and reported with mean and standard deviation across these three runs.

\begin{figure}[htbp]
    \centering
    \includegraphics[width=\linewidth]{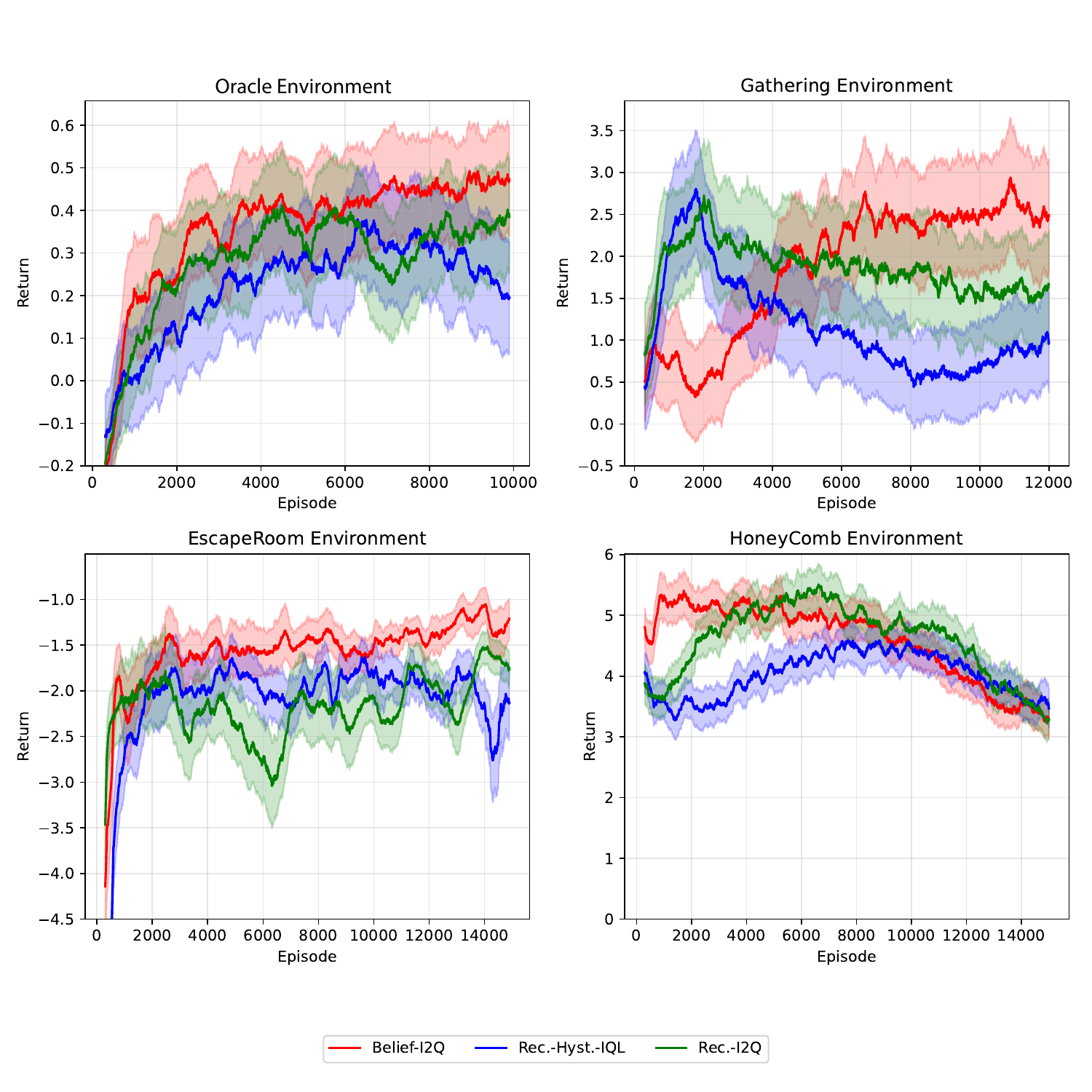}
    \caption{Evaluation results of our approach \textit{Belief-I2Q} against recurrent baselines of I2Q and hysteretic Q-learning. The plots show returns per episode, smoothed over 100 episodes. The results are averaged over three random seeds. The shaded areas show the standard deviation of the results across random seeds.}
    \label{fig:results}
\end{figure}

Our approach, Belief-I2Q, learns strong policies in all domains, apart from the HoneyComb environment, typically outperforming the benchmark solutions.
Both Rec.-I2Q and Rec.-Hyst.-IQL often struggle to learn coherent policies and exhibit conversion problems in some of the domains.
In particular, Belief-I2Q achieves stronger final performance in the Oracle, Gathering, and Escape Room environments.
We also observe faster convergence speed, for the Oracle and Escape Room environments.
While Belief-I2Q achieves better final performance in the Gathering environment, convergence is slower than for the baselines.
We speculate that this might be due to the comparatively complex belief state in this environment, which needs to capture all areas of the map, currently not in any agents' radius.
This, in turn, could initially lead to high variance in the value function estimates before the agent has learned to interpret the belief states correctly.

The evaluated settings are chosen to reflect different nuances of partial observability.
Belief-I2Q performs particularly well in settings where all agents can infer the underlying system state from current or past observations without relying on other agents' learned policies for this representation learning aspect.
For the belief model to infer useful information, the unobserved parts of the state need to be reasonably predictable from the observation history.
This appears to be possible in the Oracle, Gathering, and Escape room environments.
We conjecture that the ability to capture uncertainty over the inferred states then enables the agents to learn better policies.
The benchmarks do not have this ability and might misguide the agent in cases where the unobserved part of the state can not yet be accurately predicted.
In contrast to the three mentioned environments, the HoneyComb environment introduces information asymmetry, where only two agents have access to the unobserved elements of the underlying state.
The uninformed agents would, therefore, only be able to infer the location of the higher reward field from the movement patterns of other (informed) agents.
As the belief model is trained using random rollouts, however, the movement patterns are likely not reflective of this information and no leader-follower dynamic can be established.
In our results, we consequently observe comparatively poor performance in this environment.
The chosen benchmarks also do not perform well in this scenario, indicating that they also fail to infer information about the underlying state from the movement patterns of informed agents to bridge this information asymmetry.

\begin{figure}[htbp]
    \begin{subfigure}[t]{0.45\textwidth}
        \includegraphics[width=\textwidth]{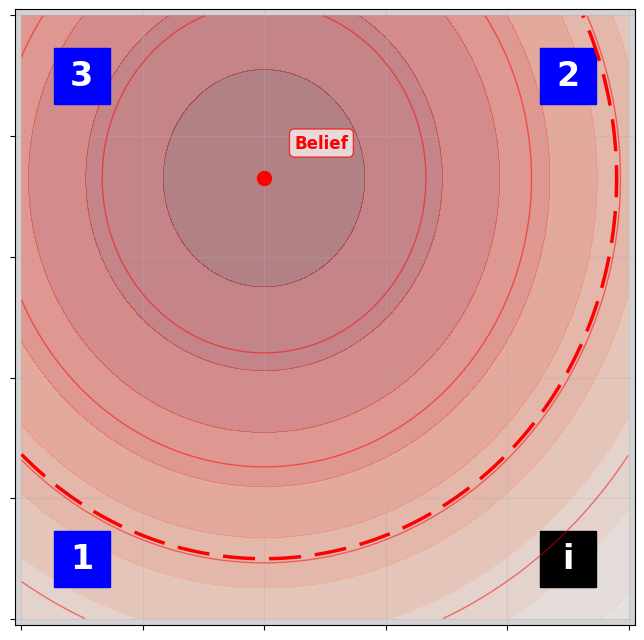}
        \caption{Belief state one time step before querying the oracle. The belief states are not grouped in this case, as the oracle has not been queried yet, and no information on the correct box location is available.}
        \label{fig:belief_before}
    \end{subfigure}
    \hfill
    \begin{subfigure}[t]{0.45\textwidth}
        \includegraphics[width=\textwidth]{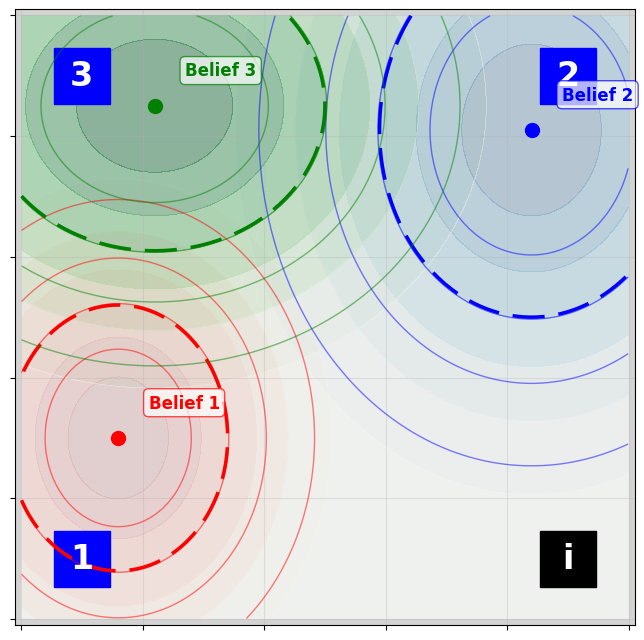}
        \caption{Belief states one time step after querying the oracle. The belief states are grouped according to where the correct box was actually located~--~Belief 1 corresponds to episodes where the correct box was in the bottom-left corner (Box 1) and so forth.}
        \label{fig:belief_after}
    \end{subfigure}
    \caption{Visualization of the belief state before (LHS) and after (RHS) querying the oracle.
    The output from the belief model, mean and standard deviation are averaged over 100 episodes per belief state visualization.
    The plotted standard deviation is, therefore, the average of standard deviations from the belief model across samples, not the standard deviation of means across the samples. 
    The dashed lines in the contours represent one standard deviation.
    For this visualization, we only query belief states from one agent.}
    \label{fig:belief_state_viz}
\end{figure}

To evaluate the effectiveness of our belief model, beyond improving the policy performance of the RL agents, we conduct a visual inspection of the belief states over the unobserved parts of the underlying system state using the Oracle environment as an example.
We choose this environment as the information necessary to form an accurate belief state is revealed at a particular moment in time, i.e., when querying the oracle, allowing an easy before and after comparison of the belief states.
Such a comparison is not easily accomplished for the other environments since the belief evolves continuously as the agents explore their surroundings.
Additionally, the belief states are easily interpretable in this case, as the true location of the reward is revealed and the belief states can be grouped according to this information.
We present this visualization of belief states in Figure \ref{fig:belief_state_viz}.
Before querying the oracle (Figure \ref{fig:belief_before}), the belief states show high variance and do not point towards any particular treasure location as being more likely.
Note that the belief still approximately captures the average location of the boxes, i.e., the mean is approximately at the mid-point of the treasure locations.
After querying the oracle (Figure \ref{fig:belief_after}), we observe that the belief states clearly differentiate between the three possible treasure locations.
It should be noted that the belief states for this comparison are collected several steps into the episode, i.e., when one of the agents has moved to the oracle to query it.
The displayed belief states could, therefore, be impacted by any exploration already done by the other agent, which might have moved to one of the treasure locations, ruling it out or confirming it as the correct box.
However, this visualization still provides evidence that Belief-I2Q can effectively capture and represent the partially observable aspects of the environment.

\section{Conclusion}

We present a new approach for multi-agent RL under partial observability and evaluate it in several experimental settings exhibiting different types of partial observability.
Our experimental results reveal the effectiveness of the proposed approach when compared with relevant algorithms in the field.
By leveraging system state information in the initial pre-training phase, we are able to fully decentralize training and execution during the reinforcement learning stage.
Our algorithm could also be easily combined with existing policy gradient methods by using the learned Q-function as a critic.
Future research could explore this avenue and extend applications to larger-scale problems.

\bibliographystyle{plainnat}
\bibliography{references}




\appendix
\section{Model Architecture}
\label{app:architecture}

\subsection{Belief-I2Q}
The implementation consists of several elements, all parametrized by neural networks.
The history encoder shared between $q_{\phi}$ and $p_{\theta}$, is implemented as a single-layer GRU with $[64]$ hidden units.
The encoder and decoder networks $q_{\phi}$ and $p_{\theta}$ are both implemented as 2-layer feed-forward MLPs with sizes $[64, 64]$ using ReLu activations.
The encoder outputs $z_i$ of size belief\_dim, while the decoder outputs the belief over unobserved state features and log-variance, the dimensionality of which is dictated by the environment.
Both $z_i$ and the state predictions $\hat{s}$ are modelled as Gaussian distributions.
The $Q_{ss}$, $f$ and $Q$ functions are parameterized by 3-layer feed-forward neural networks using ReLu activations with hidden sizes $[128,128,128]$. 
We also use target networks for $Q_{ss}$ and $Q$, which are updated $\tau = 0.005$ every episode.
We use Adam optimizers for all components.

\subsection{Baselines}
For recurrent hysteretic Q-learning, we parameterize the Q-function by a single-layer GRU with $[64]$ hidden units, followed by three linear layers using ReLu activations and $[128, 128, 128]$ hidden units.
We use a target network for $Q$, updated with $\tau = 0.005$ every episode and use an Adam optimizer.

The recurrent I2Q implementation follows the one used by \cite{jiang2022i2q}.
We use a single-layer GRU with $[64]$ hidden units, shared between the $Q^{ss}$ and $Q$ function.
The $Q_{ss}$, $f$ and $Q$ functions are parameterized by 3-layer feed-forward neural networks using ReLu activations with hidden sizes $[128,128,128]$. 
Again, we use target networks for $Q_{ss}$ and $Q$, which are updated $\tau = 0.005$ every episode, and train all components using Adam optimizers.

All models are implemented using PyTorch \cite{paszke2017pytorch}.

\section{Hyperparameters}
\label{app:hyperparameters}
All models are trained using a replay buffer with a capacity of $10,000$ episodes, an epsilon-greedy policy based on the learned Q-values with a linearly decaying epsilon, starting at $\epsilon = 0.6$, and target networks, which are updated after every episode with $\tau=0.005$.
We limit the maximum episode length to $40$, $100$, $100$, and $25$ steps for the Oracle, Gathering, EscapeRoom and HoneyComb environments respectively.
The baselines are trained using a batch size of $32$ episodes with one update step per episode. Since Belief-I2Q is trained using individual transitions, we adjust the update batch size to reflect this.
We conduct a grid search for all models. 
We search the learning rates for the $Q$, $Q^{ss}$, the belief model and $f$ over $\{0.001, 0.0003\}$.
We also search the $\lambda$ parameter for $f$ over $\{0.1, 0.3\}$ and the latent dimension of $z_i$ over $\{8, 16, 32\}$.
The reported results reflect the runs that performed best, averaged over 3 random seeds.


\end{document}